\documentclass[runningheads]{llncs}

\usepackage{eccv}

\usepackage{eccvabbrv}

\usepackage{graphicx}
\usepackage{booktabs}

\usepackage[accsupp]{axessibility}  

\usepackage{threeparttable}
\usepackage{caption}
\usepackage{hyperref}

\usepackage{orcidlink}

\begin{document}

\title{Rethinking Features-Fused-Pyramid-Neck for Object Detection} 
\author{Hulin Li\inst{}\orcidlink{0000-0002-0266-8720}}
\authorrunning{Hulin Li}
\institute{College of Traffic and Transportation, Chongqing Jiaotong University, Chongqing 400074, China\\
\email{alan@mails.cqjtu.edu.cn}}

\maketitle
\begin{abstract}
Multi-head detectors typically employ a features-fused-pyramid-neck for multi-scale detection and are widely adopted in the industry. However, this approach faces feature misalignment when representations from different hierarchical levels of the feature pyramid are forcibly fused point-to-point. To address this issue, we designed an independent hierarchy pyramid (IHP) architecture to evaluate the effectiveness of the features-unfused-pyramid-neck for multi-head detectors. Subsequently, we introduced soft nearest neighbor interpolation (SNI) with a weight-downscaling factor to mitigate the impact of feature fusion at different hierarchies while preserving key textures. Furthermore, we present a features adaptive selection method for downsampling in extended spatial windows (ESD) to retain spatial features and enhance lightweight convolutional techniques (GSConvE). These advancements culminate in our secondary features alignment solution (SA) for real-time detection, achieving state-of-the-art results on Pascal VOC and MS COCO. Code will be released at \href{https://github.com/AlanLi1997/rethinking-fpn}{https://github.com/AlanLi1997/rethinking-fpn}. This paper has been accepted by \href{https://link.springer.com/chapter/10.1007/978-3-031-72855-6_5}{\textbf{ECCV2024}} and published on Springer Nature.
\end{abstract}
\section{Introduction}
\label{sec:intro}
\begin{figure}[h]
		\begin{center}
	 	\includegraphics[width=0.8\linewidth]{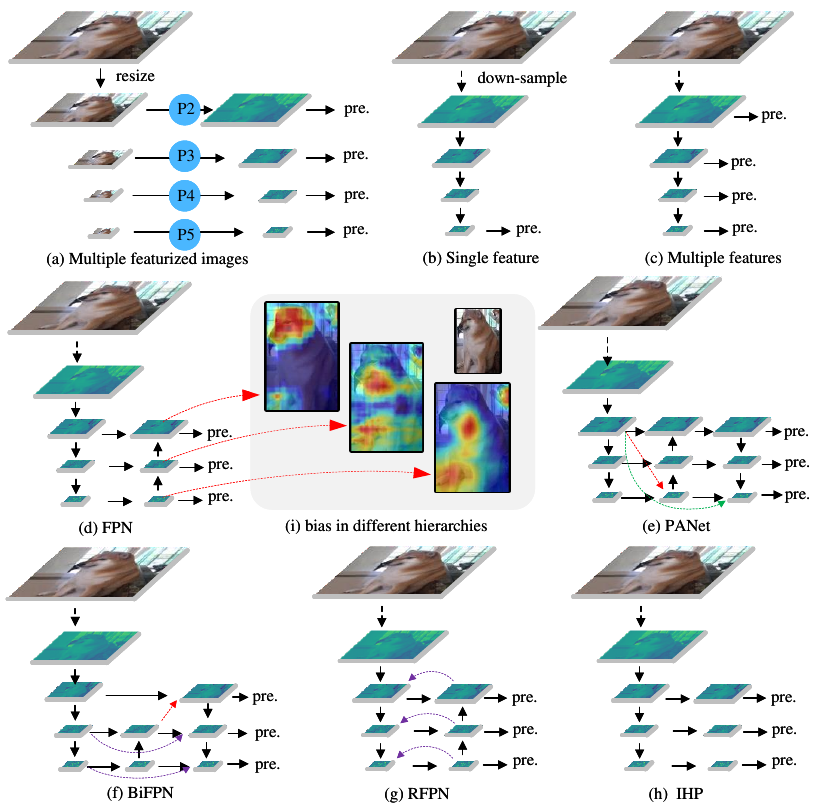}
		\end{center}
		\caption{The IHP and seven other typical neck architectures. The three heat maps are come from the third (P3), fourth (P4), and fifth (P5) hierarchies of the FPN-neck model, respectively. The area of interest for each hierarchies' features is highlighted in red; the darker the color, the higher the feature weight. It is evident that lower-level hierarchies prioritize local features, while higher-level hierarchies focus more on global features. In other words, there is a representation bias at different hierarchical levels. If these features are combined by point-to-point fusion without first addressing the misalignment issue, it may introduce noise rather than enrich semantics.}
		\label{f1}
\end{figure}
Bounding box regression for object detection and pixel-level boundary regression for instance segmentation are pivotal and challenging visual tasks. Unlike basic image classification\cite{imagenet} , detection or segmentation tasks often involve scenes with multiple objects of varying scales. Early deep learning-based object detection models began exploring real-time multi-scale object detection. RCNN \cite{rcnn} employed selective search to extract a large number of region proposals ($\sim$ 2k). Although the vast number of image patches (region proposals) might aid multi-scale detection, in practice, most scenes do not contain such a high number of objects to recognize. This approach was later deemed inefficient \cite{frcnn}, as the proposed regions were derived from the image rather than features, necessitating a convolutional neural network (CNN) to extract features and predict classes for each proposal. A simplified RCNN architecture, as depicted in \cref{f1} (a), converts images of different scales into feature maps for object detection, largely neglecting the real-time detection capability of models. YOLO \cite{yolo} introduced the first real-time deep learning-based object detection model, which did not specifically focus on multi-scale detection, with a simplified structure shown in \cref{f1} (b). SSD \cite{ssd} effectively combined the strengths of RCNN and YOLO, emphasizing the concept of multi-scale detection. It was the first model to achieve real-time multi-scale object detection using multi-scale features, with an architecture similar to \cref{f1} (c). SSD efficiently utilized feature maps of different scales from the VGG-16 backbone\cite{vgg} to predict objects of varying scales. The exploration by early deep learning-based object detection models provided an important insight to the field: the efficient utilization of multi-scale features is crucial for enhancing the overall performance (accuracy-speed trade-off) of detectors. Consequently, FPN \cite{fpn} introduced the feature pyramid-based multi-scale feature fusion concept, which gradually became a structural paradigm for detectors. This led to numerous representative works enhancing multi-scale feature fusion, such as PANet \cite{panet}, ASFF \cite{asff}, BiFPN \cite{bifpn}, and RFPN \cite{rfpn}. The performance of real-time detectors has improved with the implementation of feature fusion strategies. However, further advancements are hindered by the increasing complexity of these techniques. Before the introduction of FPN, object detectors primarily relied on structures without fusion\cite{rcnn,yolo,ssd}, as depicted in \cref{f1} (a)-(c). With the advent of FPN, these detectors have shifted towards FPN-based paradigms or more intricate improvement methods\cite{frcnn,yolo,yolo9000,yolov3,fssd}, as illustrated in \cref{f1} (e)-(g). It is important to note that the FPN-like strategy should not be considered an obligatory component for all detectors. Representation learning suggests that models ultimately require refined representations rather than an abundance of complex representations. Under the FPN-like paradigm, feature maps of different hierarchies and  different resolutions are combined at the element level (point-to-point), necessitating the expansion of low-resolution, high-level feature maps to match the resolution of low-level feature maps. However, there are inherent representation biases in feature maps at different hierarchical levels: low-level feature maps tend to retain representations of small objects (local features in small receptive fields), while high-level feature maps favor representations of large objects (global features in large receptive fields), as shown in \cref{f1} (i). The direct element-level fusion of these feature maps with representation bias may lead to partial destruction of representations and exacerbate the feature misalignment problem during fusion. 

In visual models, the resolution of feature maps decreases progressively from low to high hierarchical levels of the feature pyramid, achieved through step-by-step downsampling. Typically, downsampling reduces the resolution (width × height) of feature maps at a square root rate, inherently filtering out some spatial information regardless of whether the number of features (channels) remains constant or increases. It is also noteworthy that the transition of feature information from spatial to channel dimensions primarily occurs during downsampling, underscoring the importance of this stage. Traditional downsampling methods operate in a single mode, such as convolution with a stride of 2 (learnable) or average pooling and max pooling with a stride of 2 (non-learnable). These single-mode approaches have persisted since the introduction of ResNet\cite{resnet} without significant evolution, which merits further investigation. Additionally, the initial purpose of FPN was not only to demonstrate the advantages of feature fusion in addressing multi-scale detection challenges but also to appropriately reduce model complexity. In industries with limited hardware resources, complex models are often impractical\cite{xcption}\cite{ghostnet}. Therefore, we enhance the lightweight convolution technique of the GSConv\cite{gsconv} to achieve a preferable trade-off between accuracy and speed for real-time detection models.

The main contributions of this work can be summarized as follows:

1).We rethink the effectiveness of FPN-like paradigms for modern real-time detectors based on representation learning and identify issues with feature misalignment during element-level fusion.

2).We design an independent hierarchy pyramid architecture (IHP) without feature fusion in the neck to validate our findings. The IHP achieves advanced results.

3).We introduce soft nearest neighbor interpolation (SNI) for up-sampling to mitigate feature misalignment during fusion.

4).We provide a feature adaptive selection method in extended spatial windows for downsampling (ESD) to enhance spatial feature retention during the downsampling stage. This method is plug-and-play and optimizes performance at a low cost.

5).We simplify and enhance the GSConv lightweight convolution (GSConvE) to offer more cost-effective options for models operating on edge devices with resource constraints.

6).We validate our approaches on Pascal VOC and COCO and present the SA solution. Existing lightweight real-time detection models can be optimized with the SA solution without bells and whistles.

\section{Related Work}
\subsection{Multi-head detection and features fusion}
The first-generation general detection models using deep learning consist of two main components: the backbone and the detecting-head \cite{rcnn, yolo}. These models typically use final feature maps for prediction. SSD \cite{ssd} introduced the use of multi-level feature maps for object detection, marking the beginning of multi-head detection and influencing subsequent real-time detector designs \cite{yolov3}. In the SSD model, each detecting-head directly uses raw feature maps from different levels of the backbone without any fusion. The key advantage of the multi-head detection method is the ability to predict objects of different scales using feature maps with varying receptive fields. High-level feature maps, with their wide global receptive fields, are beneficial for identifying large objects, while low-level feature maps, with refined local receptive fields, are helpful for identifying small objects. FPN introduced a fusion scheme for multi-level feature maps to enhance detection accuracy, setting the trend for feature fusion and becoming a common practice. This led to the development of Backbone-Neck-Head architectures. More complex variants based on FPN, such as BiFPN \cite{bifpn} and PANet \cite{panet}, have since been proposed, enabling the use of more features for prediction. However, as prediction accuracy improved, networks became increasingly complex due to additional fusion layers. Some approaches have attempted to bypass feature fusion schemes by exploring new perspectives on multi-scale detection challenges, such as optimizing multi-scale training strategies \cite{snip} or using dilated convolutions to capture multi-scale receptive fields \cite{tnet}. But these methods have gradually lost their competitiveness with the continuous optimization of classification-backbone feature extraction capabilities and the ongoing development of feature fusion techniques \cite{yolov4}.
\subsection{Lightweight}
The choice between cloud computing or edge computing for a model depends on its number of parameters and floating-point operations (FLOPs). Therefore, reducing the number of parameters or FLOPs is a primary focus in lightweight model studies. Direct lightweight approaches include reducing the depth (number of layers) or width (number of neurons/filters) of the model, as seen in models like YOLO-fast/tiny/nano \cite{yolov3, yolov4, yolov5, yolov6, yolov7, yolov8}. Additionally, sparse computing techniques, such as depth-wise separable convolution used in MobileNets \cite{mobilenets} and ShuffleNet \cite{shufflenet}, can also be employed. However, low-depth networks often suffer from underfitting due to insufficient nonlinear representation capabilities. Therefore, reducing the number of network layers may not always be a cost-effective way to lighten the model.
\section{Secondary Features Alignment Solution}
Key contents of this work are described in detail in this section. Specifically, there are the IHP structure to directly demonstrate features unalignment problem by abandoning fusion, the SNI to alleviate unaligned during features fusion by point-to-point, the ESD to enhance spatial-features capture during the downsampling stage, and the GSConvE to improve the performance between the accuracy and speed of lightweight models.
\subsection{Independent hierarchy pyramid architecture}
The introduction of FPN brought all real-time detectors into the realm of feature fusion. However, we have identified a significant issue arising from the point-to-point fusion of different level features – partial local features become unaligned. This fusion process is akin to adding noise when unaligned features are combined, as features unrelated to the target space are forcibly integrated, resulting in spatial disarray. For instance, as depicted in \cref{f2}, adding the head-feature of a puppy to the butt creates an undefined breed when fed to the detection-head. This scenario is particularly evident when the puppy occupies fewer pixels. Typically, nearest neighbor interpolation is used to rapidly increase the resolution of high-level feature maps to match that of low-level feature maps, followed by fusion through methods such as element-wise addition, channel concatenation, or weighted sum. However, downsampling tends to be nonlinear and irreversible, leading to inconsistencies in spatial features among the additional features generated by up-sampling. Although BiFPN \cite{bifpn} offers a learnable fusion method, it has not been widely adopted by recent state-of-the-art real-time detectors, such as YOLOv7 \cite{yolov7} or YOLOv8 \cite{yolov8}, due to the increased complexity and lack of demonstrated comprehensive performance improvements in these models.

The IHP adopts a radical approach by abandoning all fusions in the neck, fundamentally circumventing the features misalignment problem and streamlining the structure. By leveraging the inherent benefits of the multi-head detection mode, the IHP directly predicts objects of different sizes using feature maps of different levels. It is important to note that there are significant differences between IHP and the SSD architecture: SSD directly uses different level features from the backbone to detect objects but the IHP introduces a bottleneck convolution module before prediction to filter the features. The advantage of this approach is that it allows the model to independently learn features for a specific scale branch without affecting features at other levels. Directly stacking convolutional blocks at different levels of the backbone can lead to interference between different levels, which we believe might be the primary reason why using a very deep backbone in object detection models can result in performance degradation. For instance, DSSD\cite{dssd} showed a performance drop when ResNet-101 was used as the backbone instead of VGG-16 in the SSD detector. In Part \uppercase\expandafter{\romannumeral1} of \cref{t1}, all baselines of coupled-head YOLOs demonstrated improved results when using the IHP. Features fusion often overlooks the potential introduction of noise when augmenting semantic richness for prediction. While the IHP may appear to reduce semantic information, it effectively harnesses the advantages of multi-head detection to circumvent the misalignment problem. This strategic utilization of multi-head detection is the key factor behind the competitive results achieved by the IHP. Unfortunately, unlike localization-classification-coupled-head detectors\cite{yolo,yolo9000,yolov3,yolov5,yolov7}, the IHP did not achieve the same positive reaction to localization-classification-decoupled-head detectors\cite{yolov6,yolov8}. Decoupled-head detectors address a key challenge faced by coupled-head detectors: the disparity in the interests of classification and localization tasks, both of which are often performed using the same layer for prediction \cite{rethink}. This approach is akin to forcibly fusing misaligned features for prediction. Decoupled-head detectors overcome this limitation by independently predicting classification and localization tasks through separate branches. This enables them to learn beneficial representations from redundant features, which is not possible in coupled-head detectors. Consequently, while FPN strategies are still applicable to some extent in decoupled-head detectors, the features misalignment problem still needs to be addressed.
\subsection{Soft nearest neighbor interpolation}
\begin{figure}[h]
		\begin{center}
		\includegraphics[width=1.0\linewidth]{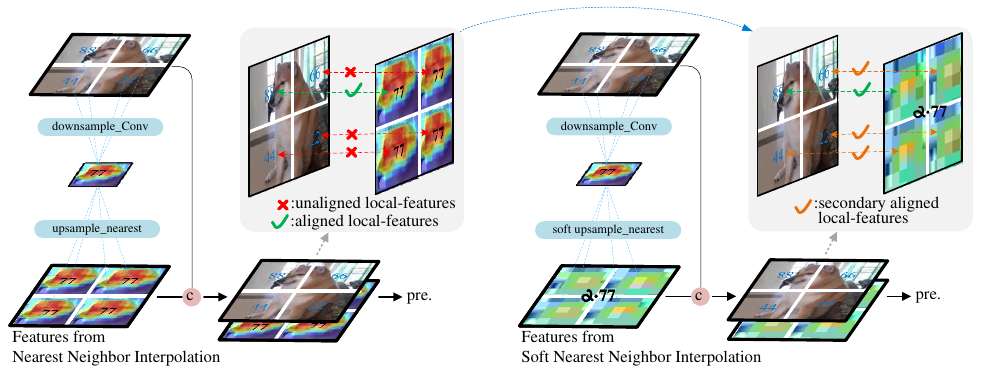}
		\end{center}
		\caption{Illustrations of the features misalignment in fusion and the SNI. These numbers, 22-88, are just markers of different local-features not real feature values.}
		\label{f2}
\end{figure}
\begin{figure}[th]
  \centering
  \includegraphics[width=0.8\linewidth]{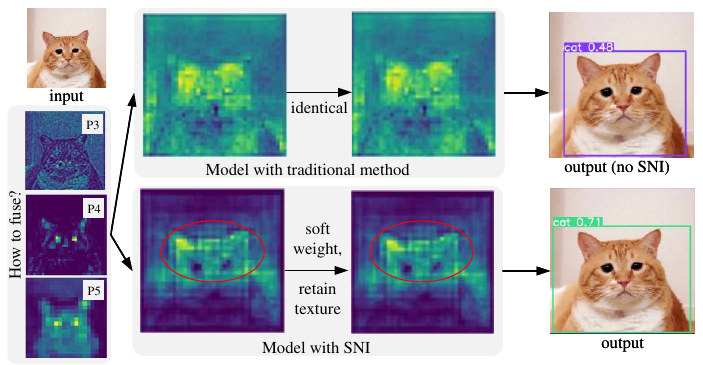}
   \caption{A comparison of the SNI and traditional method. This result could be reproduced by using SNI source code (model:Yolov5n-panet). The same size scaling before and after up-sampling is for intuitive comparison.}
   \label{sniex}
\end{figure}
In detection or segmentation models, different-level feature fusion typically occurs after the up-sampling of feature maps, with nearest neighbor interpolation and transposed convolution being popular up-sampling methods. However, the resolution expansion resulting from up-sampling in detection or segmentation differs significantly from tasks such as generation or super-resolution. In detection, the expanded feature maps represent abstract high-level semantic features rather than raw image detail information. Transposed convolution introduces increased computation and latency. To address the features misalignment problem with minimal cost while maintaining the speed of nearest neighbor interpolation, we explore a solution. Nearest neighbor interpolation is akin to average unpooling but is considered a "hard" operation. By softening this operation, similar to the transformation of SoftMax to the Max function, we aim to mitigate the "hard" features misalignment issue. To achieve this, we introduce a soft factor, denoted as $\alpha$, for the nearest neighbor interpolation during up-sampling:
\begin{equation}
  \textbf{\emph{Y}}=\alpha\cdot \textbf{\emph{f}}(\textbf{\emph{X}}),\alpha=\frac{\emph{Resolution}\textbf{\emph{X}}}{\emph{Resolution}\textbf{\emph{Y}}}
  \label{eq1}
\end{equation}
In \cref{eq1}, the resolution of the high-level feature maps (\textbf{\emph{X}}) and low-level feature maps (\textbf{\emph{Y}}) are denoted as \emph{Resolution}\textbf{\emph{X}} and \emph{Resolution}\textbf{\emph{Y}}, respectively. The nearest neighbor interpolation operation, represented by \textbf{\emph{f}}, is illustrated in \cref{f2}. The SNI adjusts the influence of high-level semantic features on low-level features based on the zoom factor for feature maps. Specifically, as the zoom factor increases, the impact of high-level semantic features on low-level features weakens. Unlike SoftMax, the SNI does not convert outputs into probabilities but rather mitigates misalignment without additional cost when integrating high-level features with low-level features through point-to-point fusion. SNI softens feature weights while preserving key textures, thereby reducing the mutual influence of features from different hierarchical levels during fusion, as qualitatively demonstrated in \cref{sniex}. This approach achieves the secondary (auxiliary) features alignment (SA) and lowers the complexity for the model to learn from the fused representations. In contrast, traditional methods that directly merge features from different hierarchical levels can cause unstable competitive effects due to representation biases, increasing the learning difficulty for the model. For example, if the three heatmaps in \cref{f1} (i) are directly fused, should the red areas cover the entire image? Clearly, the answer is no. If such a situation occurs, it indicates a failure in prior feature extraction, as the objective of a model is to capture and focus on useful features. This is also why models under the FPN-like paradigm typically require several additional convolutional blocks to further filter features after fusing and before predicting.

In Part \uppercase\expandafter{\romannumeral2} of \cref{t1}, all tested state-of-the-art models demonstrate accuracy improvements solely by incorporating the SNI, confirming the presence of features misalignment in FPN-based models. Notably, for YOLOv6 \cite{yolov6}, a detector equipped with various techniques tailored for industry applications, the AP is enhanced by up to 2 percentage points just with the adoption of SNI, while reducing computational cost and inference latency. We achieved significant optimization results in the original YOLOv6 by simply replacing transposed convolution with SNI, highlighting the efficacy of this approach.
\subsection{Features adaptive selection in extended spatial windows}
\begin{figure}[h]
		\begin{center}
		\includegraphics[width=1.0\linewidth]{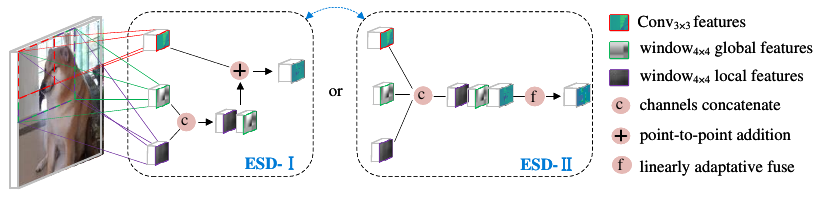}
		\end{center}
		\caption{Illustrations of the ESD-\uppercase\expandafter{\romannumeral1} and ESD-\uppercase\expandafter{\romannumeral2}. It must be mentioned that although the size of the extended window does not have to be same in the ESD-\uppercase\expandafter{\romannumeral1}, pooling is a non-learnable operation, so the features from an unequal window may become noise for other and bring the trouble of the features misalignment again. The ESD-\uppercase\expandafter{\romannumeral2} is free on size of the extended window because of the the learnable fusion like the SPPNet \cite{sppnet} for the YOLOv3.}
		\label{f3}
\end{figure}
Feature downsampling is a critical technique in vision models based on deep learning, involving the progressive reduction of feature map resolution. In vision representation learning, models typically rely on low-resolution, high-level feature maps for prediction rather than directly processing high-resolution raw images. However, excessive downsampling can result in significant loss of spatial details \cite{resnet}. Recent detectors typically limit the number of downsampling operations to no more than five times, leading to a spatial resolution reduction of $2^5$ times, as demonstrated in YOLOs \cite{yolov3, yolov4, yolov5, yolox, ppyoloe, yolov6, yolov7, yolov8} and Swin Transformer \cite{swin}. Common downsampling methods include convolution and pooling, with models typically employing only one of these methods. We propose the features adaptive selection method in extended spatial windows for downsampling (ESD) to enhance spatial information retention capabilities. ESD consists of two nonlinear branches and one linear branch. In the nonlinear branches, a norm convolutional layer and an extended window max-pooling layer enhance local feature capturing, while extended window average-pooling enhances global feature capturing. Subsequently, linear features are merged with nonlinear features. Two fusion modes, ESD-\uppercase\expandafter{\romannumeral1} and ESD-\uppercase\expandafter{\romannumeral2}, are designed for lightweight and norm models, respectively. The difference lies in the feature fusion stage, as illustrated in \cref{f3}. ESD-\uppercase\expandafter{\romannumeral1} employs element-wise addition for merging features, offering low computational cost suitable for lightweight models, while ESD-\uppercase\expandafter{\romannumeral2} utilizes learnable linearly adaptive fusion, slightly increasing computational cost but suitable for norm models. Importantly, local and global features of the extended window are sampled using simple pooling techniques, preserving input information to a significant extent. This reduces information loss during downsampling and enables subsequent layers to learn partial representations from the previous layer, akin to indirectly realizing a hidden shortcut connection reminiscent of ResNet \cite{resnet}.
\subsection{Lightweight GSConv enhancement}
\begin{figure}[h]
		\begin{center}
	 	\includegraphics[width=0.9\linewidth]{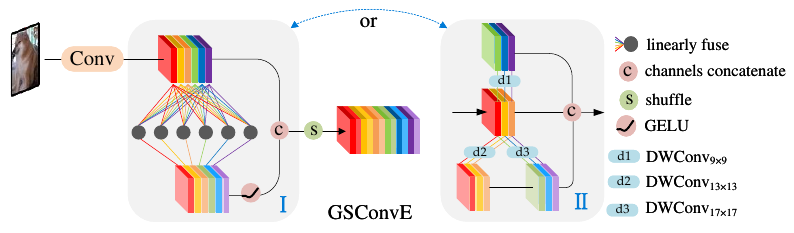}
		\end{center}
		\caption{Illustrations of the GSConvE-\uppercase\expandafter{\romannumeral1} and \uppercase\expandafter{\romannumeral2}.}
		\label{f4}
\end{figure}
GSConv \cite{gsconv} was introduced to address the channel interaction limitations of depthwise separable convolution by blending features from both vanilla convolution and depthwise convolution. The original GSConv consists of three fundamental components: a 3×3 vanilla convolution, a 5×5 depthwise convolution, and a feature shuffle operation. Building upon this, we propose two evolutions of GSConv, namely GSConvE-\uppercase\expandafter{\romannumeral1} and GSConvE-\uppercase\expandafter{\romannumeral2}, tailored for norm models and lightweight models, respectively. GSConvE-\uppercase\expandafter{\romannumeral1} is designed for norm models, where intermediate feature maps on the auxiliary branch are derived from dense linear mappings, while output feature maps originate from sparse linear mappings. This approach maintains feature richness without significantly increasing computational cost. On the other hand, GSConvE-\uppercase\expandafter{\romannumeral2} is geared towards lightweight models. It incorporates three depthwise convolution auxiliary branches with kernel sizes of 9×9, 13×13, and 17×17, respectively. By utilizing larger kernel sizes, this variant can directly capture larger receptive fields and global features with fewer layer accumulations, offering high cost-effectiveness for lightweight models that prioritize reduced FLOPs and layer count. In GSConvE-\uppercase\expandafter{\romannumeral1}, we explore the removal of the batch normalization layer on the depthwise convolution branch while retaining the batch normalization layer on the main branch, similar to ConvNeXt \cite{convnext}. This adjustment avoids the vanishing gradient problem and simplifies computation, leading to improved prediction accuracy.
\section{Experiments}
Real-time detectors are highly favored in industry due to their hardware-friendly nature and the optimal balance they strike between accuracy and speed. In our study, we selected popular YOLO models—YOLOv3 \cite{yolov3}, YOLOv4 \cite{yolov4}, YOLOv5 \cite{yolov5}, YOLOv6 \cite{yolov6}, YOLOv7 \cite{yolov7}, and YOLOv8 \cite{yolov8}—as baselines to evaluate the efficacy of the SA solution in real-time detection scenarios (SYOLO).
\subsection{Datasets}
We conducted trials on the Pascal VOC dataset \cite{voc} and the MS COCO dataset \cite{coco}. The Pascal VOC 07+12 dataset was utilized for ablation experiments, wherein models underwent training on the train\&val07+12 set and were subsequently evaluated on the test07 set. On the other hand, the MS COCO dataset was employed for comparative experiments, with models trained on the train2017 set and evaluated on the val2017 set. To ensure fairness and consistency, all released models were trained using the default set of hyperparameters and without utilizing pre-trained weights. More details of hyperparameters are in the support material.
\subsection{Ablation studies}
In \cref{t1}, we present the results of ablation experiments conducted on six baselines \cite{yolov3, yolov4, yolov5, yolov6, yolov7, yolov8}, focusing on the impact of the IHP, SNI, ESD. The lightweight experiment results of GSConvE are reported in \cref{tgsc}. All baselines retained their original neck architecture, as we believe the competitive neck has been carefully chosen in the original works. Performance metrics highlighted in green indicate performance gains, while those in red indicate losses. It is crucial to address an often-overlooked aspect during the verification process: the inconsistency between reported evaluation accuracy and actual usage. Frequently, detectors are evaluated with a very low confidence threshold, such as 0.001, which results in a large number of prediction boxes and inflated evaluation accuracy. However, in real-world detection scenarios, the confidence threshold is typically set much higher, around 0.25 or above. This filtering process ensures that higher-quality bounding boxes are utilized for practical purposes. The discrepancy between evaluation settings and real-world usage can create a misleading impression of accuracy. Therefore, it is essential to maintain consistency in configuration between evaluation and application. While this may result in a significant drop in evaluation accuracy, as shown in \cref{fc}, it provides a more realistic assessment of model performance.
\begin{table}[h]
\scriptsize
  \renewcommand{\arraystretch}{1.0}
  \centering
 \caption{Ablation experiments of the IHP, SNI and ESD-\uppercase\expandafter{\romannumeral1}/\uppercase\expandafter{\romannumeral2} on VOC 07+12.}
  \label{t1}
  \begin{tabular}{@{}l|c|c|c|c|c|c@{}}
	\toprule
 	\textbf{Models}&	\textbf{Size}&	\textbf{Param.}(M)&	\textbf{FLOPs}(G)&	${\textbf{AP}}_{50}$(\%)&	\textbf{AP}(\%)& $\textbf{Latency}_{b=16}^{T4}$(ms) \\
	\midrule	
	\multicolumn{7}{c}{\textbf{baselines}}  \\
	\midrule
	Yolov3-t\cite{yolov3}&		640&		8.71&		13.0&		58.0&		27.3&		3.8\\
	Yolov4-t\cite{yolov4}&		640&		5.91&		16.1&		62.6& 	32.2& 	4.4\\
	Yolov5-n\cite{yolov5}&		640&		1.79&		4.2&		74.0& 	47.0& 	3.4\\
	Yolov6-n\cite{yolov6}&		640&		4.30& 	11.1& 	80.2& 	58.8& 	3.5\\
	Yolov7-t\cite{yolov7}&		640&		6.06&		13.3&		78.4&		54.6&		5.0\\
	Yolov8-n\cite{yolov8}&		640&		3.01&		8.1& 		79.8&		59.7& 	3.7\\
	\midrule
	\multicolumn{7}{c}{\textbf{Part \uppercase\expandafter{\romannumeral1}: IHP} (ours)}  \\
	\midrule
	Yolov3-t-ihp&	640&		9.33&		15.0&		59.9\textcolor{green}{(↑1.9)}&		29.1\textcolor{green}{(↑1.8)}&		3.8\\
	Yolov4-t-ihp&	640&		6.63&		16.0&		63.8\textcolor{green}{(↑1.2)}&		33.9\textcolor{green}{(↑1.7)}&		4.1\\
	Yolov5-n-ihp&	640&		1.66&		3.9&		74.8\textcolor{green}{(↑0.8)}&		48.6\textcolor{green}{(↑1.6)}&		\textbf{3.2}\\
	Yolov6-n-ihp&	640&		4.37&		10.6&		79.4\textcolor{red}{(↓0.8)}&		57.4\textcolor{red}{(↓1.4)}&		3.7\\
	Yolov7-t-ihp&	640&		6.77&		14.5&		80.7\textcolor{green}{(↑2.3)}& 	55.3\textcolor{green}{(↑0.7)}&		4.7\\
	Yolov8-n-ihp&	640&		2.99&		8.1&		79.3\textcolor{red}{(↓0.5)}&		59.1\textcolor{red}{(↓0.6)}&		3.7\\
	\midrule
	\multicolumn{7}{c}{\textbf{Part \uppercase\expandafter{\romannumeral2}: SNI} (ours)}\\
	\midrule
	Yolov3-t-sni&	640&		8.71&		13.0&		58.6\textcolor{green}{(↑0.6)}&		28.0\textcolor{green}{(↑0.7)}&		3.8 \\
	Yolov4-t-sni&	640&		5.91&		16.1&		63.4\textcolor{green}{(↑0.8)}&		33.0\textcolor{green}{(↑0.8)}&		4.4 \\
	Yolov5-n-sni&	640&		1.79&		4.2&		74.5\textcolor{green}{(↑0.5)}&		47.7\textcolor{green}{(↑0.7)}&		3.4 \\
	Yolov6-n-sni&	640&		4.28&		11.0&		\textbf{81.9}\textcolor{green}{(↑1.7)}&		60.8\textcolor{green}{(↑2.0)}&		3.4 \\
	Yolov7-t-sni&	640&		6.06&		13.3&		81.5\textcolor{green}{(↑3.1)}&		57.9\textcolor{green}{(↑3.3)}&		5.0 \\
	Yolov8-n-sni&	640&		3.01&		8.1&		80.3\textcolor{green}{(↑0.5)}&		60.1\textcolor{green}{(↑0.4)}&		3.7 \\
	\midrule
	\multicolumn{7}{c}{\textbf{Part \uppercase\expandafter{\romannumeral3}: ESD-\uppercase\expandafter{\romannumeral1}} (ours)}\\
	\midrule
	Yolov3-t-esd-\uppercase\expandafter{\romannumeral1}&	640&		11.8&		22.4&		69.2\textcolor{green}{(↑11.2)}&	38.5\textcolor{green}{(↑11.2)}&	4.6\\
	Yolov4-t-esd-\uppercase\expandafter{\romannumeral1}&	640&		5.91&		16.1&		63.1\textcolor{green}{(↑0.7)}&		32.9\textcolor{green}{(↑0.7)}&		4.4\\
	Yolov5-n-esd-\uppercase\expandafter{\romannumeral1}&	640&		1.78&		4.2&		75.4\textcolor{green}{(↑1.4)}&		48.0\textcolor{green}{(↑1.0)}&		3.8\\
	Yolov6-n-esd-\uppercase\expandafter{\romannumeral1}&	640&		4.30&		11.1&		80.5\textcolor{green}{(↑0.3)}&		58.8&					3.4\\
	Yolov7-t-esd-\uppercase\expandafter{\romannumeral1}&	640&		6.83&		14.6&		79.2\textcolor{green}{(↑0.8)}&		55.3\textcolor{green}{(↑0.7)}&		5.2\\
	Yolov8-n-esd-\uppercase\expandafter{\romannumeral1}&	640&		3.22&		8.1&		80.1\textcolor{green}{(↑0.3)}&		59.8\textcolor{green}{(↑0.1)}&		3.8\\
	\midrule
	\multicolumn{7}{c}{\textbf{ESD-\uppercase\expandafter{\romannumeral2}} (ours)}  \\
	\midrule
	Yolov3-t-esd-\uppercase\expandafter{\romannumeral2}&	640&		8.78&		14.8&		63.0\textcolor{green}{(↑5.0)}&	32.1\textcolor{green}{(↑4.8)}&	4.9\\
	Yolov4-t-esd-\uppercase\expandafter{\romannumeral2}&	640&		5.93&		16.6&		63.9\textcolor{green}{(↑1.5)}&	33.5\textcolor{green}{(↑1.3)}&	4.7\\
	Yolov5-n-esd-\uppercase\expandafter{\romannumeral2}&	640&		2.32&		4.7&		75.7\textcolor{green}{(↑1.7)}&	48.9\textcolor{green}{(↑1.9)}&	3.6\\
	Yolov6-n-esd-\uppercase\expandafter{\romannumeral2}&	640&		4.32&		11.1&		80.4\textcolor{green}{(↑0.2)}&	58.8&				3.5\\
	Yolov7-t-esd-\uppercase\expandafter{\romannumeral2}&	640&		7.69&		15.7&		79.3\textcolor{green}{(↑0.9)}&	55.2\textcolor{green}{(↑0.6)}&	5.9\\
	Yolov8-n-esd-\uppercase\expandafter{\romannumeral2}&	640&		3.18&		8.5&		80.4\textcolor{green}{(↑0.6)}&	59.9\textcolor{green}{(↑0.2)}&	4.0\\
	\bottomrule
  \end{tabular}
\end{table}
\subsection{Comparasion experiments}
In \cref{t12} and \cref{t2}, we compare state-of-the-art real-time detectors and SYOLO (architecture in \cref{tarchi}) on the Pascal VOC and COCO benchmarks. In \cref{tsota}, we provide additional comparisons of state-of-the-art detectors on COCO \cite{coco}, including both real-time (CNNs) and non-real-time (Transformers/CNNs) models. Parameters and FLOPs metrics significantly higher than those of real-time detector models are highlighted in red.
\begin{table}[t]
\scriptsize
  \renewcommand{\arraystretch}{1.0}
  \centering
 \caption{Ablation experiments of the GSConv and GSConvE-\uppercase\expandafter{\romannumeral1}/\uppercase\expandafter{\romannumeral2} on VOC 07+12.}
  \label{tgsc}
  \begin{tabular}{@{}l|c|c|c|c|c|c@{}}
	\toprule
 	\textbf{Models}&	\textbf{Size}&	\textbf{Param.}(M)&	\textbf{FLOPs}(G)&	${\textbf{AP}}_{50}$(\%)&	\textbf{AP}(\%)& $\textbf{Latency}_{b=16}^{T4}$(ms) \\
	\midrule	
	\multicolumn{7}{c}{\textbf{baselines}(GSConv\cite{gsconv})}  \\
	\midrule
	Yolov3-tiny-gsconv&		640&		4.61&		8.3&		69.9&		41.1&		3.9\\
	Yolov4-tiny-gsconv&		640&		3.42&		12.4&		73.4&		45.2&		4.6\\
	Yolov5-n-gsconv&		640&		1.51&		3.9&		74.2&		47.9&		3.5\\
	Yolov6-n-gsconv&		640&		4.27&		11.0&		80.4&		59.1&		3.4\\
	Yolov7-tiny-gsconv&		640&		5.00&		11.4&		78.4&		55.3&		5.6\\
	Yolov8-n-gsconv&		640&		2.74&		7.8&		80.9&		61.2&		3.7\\
	\midrule
\multicolumn{7}{c}{\textbf{Part \uppercase\expandafter{\romannumeral4}: GSConvE-\uppercase\expandafter{\romannumeral1}} (ours)} \\
	\midrule
	Yolov3-t-gse-\uppercase\expandafter{\romannumeral1}&	640&		7.86&		11.6&		71.7\textcolor{green}{(↑1.8)}&			43.8\textcolor{green}{(↑2.7)}&	3.9\\
	Yolov4-t-gse-\uppercase\expandafter{\romannumeral1}&	640&		4.88&		14.3&		74.9\textcolor{green}{(↑1.5)}&			47.8\textcolor{green}{(↑2.6)}&	4.7\\
	Yolov5-n-gse-\uppercase\expandafter{\romannumeral1}&	640&		1.76&		4.2&		75.8\textcolor{green}{(↑1.6)}&			50.1\textcolor{green}{(↑2.2)}&		3.5\\
	Yolov6-n-gse-\uppercase\expandafter{\romannumeral1}&	640&		4.30&		11.1&		80.1\textcolor{red}{(↓0.3)}&			58.8\textcolor{red}{(↓0.3)}&		3.6\\
	Yolov7-t-gse-\uppercase\expandafter{\romannumeral1}&	640&		6.21&		13.5&		79.1\textcolor{green}{(↑0.7)}&			56.4\textcolor{green}{(↑1.1)}&		5.8\\
	Yolov8-n-gse-\uppercase\expandafter{\romannumeral1}&	640&		2.96&		8.1&		81.2\textcolor{green}{(↑0.3)}&			61.3\textcolor{green}{(↑0.1)}&		3.8\\
	\midrule
	\multicolumn{7}{c}{\textbf{GSConvE-\uppercase\expandafter{\romannumeral2}} (ours)}\\
	\midrule
	Yolov3-t-gse-\uppercase\expandafter{\romannumeral2}&	640&		2.85&		6.3&		70.1\textcolor{green}{(↑0.2)}&			41.4\textcolor{green}{(↑0.3)}&	4.0\\
	Yolov4-t-gse-\uppercase\expandafter{\romannumeral2}&	640&		2.36&		10.8&		75.6\textcolor{green}{(↑2.2)}&			48.0\textcolor{green}{(↑2.8)}&	4.6\\
	Yolov5-n-gse-\uppercase\expandafter{\romannumeral2}&	640&		1.48&		3.9&		76.3\textcolor{green}{(↑2.1)}&			51.4\textcolor{green}{(↑3.5)}&		3.5\\
	Yolov6-n-gse-\uppercase\expandafter{\romannumeral2}&	640&		4.28&		11.1&		80.3\textcolor{red}{(↓0.1)}&			58.8\textcolor{red}{(↓0.3)}&							3.5\\
	Yolov7-t-gse-\uppercase\expandafter{\romannumeral2}&	640&		4.33&		10.7&		79.1\textcolor{green}{(↑0.7)}&			56.2\textcolor{green}{(↑0.9)}&		5.8\\
	Yolov8-n-gse-\uppercase\expandafter{\romannumeral2}&	640&		2.67&		7.7&		79.9\textcolor{red}{(↓1.0)}&			60.4\textcolor{red}{(↓0.8)}&		3.8\\
	\bottomrule
  \end{tabular}
\end{table}
\begin{table}[t]
\scriptsize
  \renewcommand{\arraystretch}{1.0}
  \centering
  \setlength{\tabcolsep}{2mm}{
  \caption{A contrast of the different confidence threshold for the SYOLO on COCO.}
  \label{fc}
  \begin{tabular}{@{}c|c|c|c|c@{}}
	\toprule
 	\textbf{Confidence threshold}&	\textbf{AP}(\%)& ${\textbf{AP}}_{s}$(\%)& ${\textbf{AP}}_{m}$(\%)&  ${\textbf{AP}}_{l}$(\%) \\
	\midrule	
	0.001(valuation)&		53.1&				35.5&				58.7&				69.9\\
	0.25(application)&		48.7\textcolor{red}{(↓4.4)}&29.7\textcolor{red}{(↓\textbf{5.8})}&54.5\textcolor{red}{(↓4.2)}&66.5\textcolor{red}{(↓3.4)}\\
	\bottomrule
  \end{tabular}}
\end{table}

In \cref{t12}, SNI increased the AP of YOLOv7 by 0.9 percentage points without any additional computational or memory costs. SYOLO, utilizing the SA solution, achieved the highest accuracy among competitive models on the Pascal VOC benchmark. In \cref{t2}, SYOLO also achieved the highest accuracy among competitive models on the MS COCO benchmark at input resolutions of 416×416 and 640×640. Positive performance improvements across different benchmarks demonstrate the effectiveness of the SA solution for real-time object detection.
\begin{table}[thb]
  \scriptsize
  \renewcommand{\arraystretch}{1.2}
  \centering
  \setlength{\tabcolsep}{1mm}{
  \begin{threeparttable} 
  \caption{Comparasions of state-of-the-art real-time detectors on VOC 07+12.}
  \label{t12} 
  \begin{tabular}{@{}l|c|c|c|c|c|cc@{}}
	\toprule
	\textbf{Models}&	\textbf{Year}&		\textbf{Neck}&		\textbf{Size}&		\textbf{Param.}(M)&	\textbf{FLOPs}(G)&	\textbf{AP}(\%)&	$\textbf{Latency}_{b=16}^{T4}$(ms) \\
	\midrule	
	Yolov3-spp\cite{yolov3}&		2018&		FPN&		640&		61.60&		154.9&			60.5&		39.4\\
	Yolov4\cite{yolov4}&			2020&		PANet$\dagger$&	640&	52.57&		119.3&		 	66.2& 	31.0\\	
	Yolov5\cite{yolov5}&			2022&		PANet$\dagger$&		640&		46.21&		108.0&		 	63.9& 	30.1\\
	YoloX\cite{yolox}&			2021&		PANet$\dagger$&		640&		53.79&		153.7&			66.7&		40.2\\
	Yolov6\cite{yolov6}&			2022&		PANet$\dagger$&		640&		65.81& 		159.1& 		 	69.2& 	29.5\\
	Yolov7\cite{yolov7}&			2022&		PANet$\dagger$&		640&		36.58&		103.5&			68.2&		23.2\\
	Yolov8\cite{yolov8}&			2023&		PANet$\dagger$&		640&		43.62&		164.9& 			69.3& 	31.6\\
	SYolov7(ours)&				2023&		SA-PANet$\dagger$$\dagger$&	640&		36.58&	103.5&		69.1&		23.2\\
	\textbf{SYolo}(ours)&			2023&		SA&				640&		57.23&		142.4& 		\textbf{69.6}& 	29.3\\
	\bottomrule
  \end{tabular}
  \begin{tablenotes}
\scriptsize
  \item[$\dagger$] Using simplified PANet: the red and green shortcuts are cancelled in the \cref{f1} (e). 
  \item[$\dagger$$\dagger$] Using SNI replace the nearest neighbor interpolation to up-sampling in the PANet$\dagger$.
  \end{tablenotes} 
\end{threeparttable}}
\end{table}
\begin{table}[thb]
\scriptsize
  \renewcommand{\arraystretch}{1.1}
  \centering
  \setlength{\tabcolsep}{0.5mm}{
  \caption{Comparisions of state-of-the-art real-time detectors on COCO.}
  \label{t2}
  \begin{tabular}{@{}l|c|c|c|ccccc@{}}
	\toprule
 	\textbf{Models}&	\textbf{Size}&	\textbf{Param.}(M)&	\textbf{FLOPs}(G)&	\textbf{AP}(\%)& ${\textbf{AP}}_{s}$(\%)& ${\textbf{AP}}_{m}$(\%)&  ${\textbf{AP}}_{l}$(\%)&  $\textbf{Latency}_{b=16}^{T4}$(ms) \\
	\midrule	
	YoloX-t\cite{yolox}&		416&		5.1&		6.5&			32.8&		-&	-&	-&4.3\\
	\textbf{SYolo-n}(ours)&		416&		4.8&		12.0&		\textbf{35.9}& \textbf{14.3}&	\textbf{39.2}&		\textbf{54.4}&		4.3\\
	\midrule	
	YoloX-t\cite{yolox}&		640&		5.1&		15.4&			34.7&		-&-&-&7.2\\
	Yolov5-s\cite{yolov5}&		640&		7.2&		16.5&			37.4& 	-&-&-&9.8\\
	Yolov6-n\cite{yolov6}&		640&		4.7&		11.4&			37.5& 	-&-&-&4.3\\
	Yolov7-t\cite{yolov7}&		640&		6.2&		13.8&			38.7& 			18.8&		42.4&		51.9&		5.3\\
	Yolov8-n\cite{yolov8}&		640&		3.2&		8.7&			37.3& 	-&-&-&4.1\\
	\textbf{SYolo-s}(ours)&		640&		6.6&		15.7&		\textbf{40.8}& 	\textbf{21.1}&	 \textbf{45.3}&\textbf{56.9}& 		5.8\\
	\bottomrule
  \end{tabular}}
\end{table}
\begin{table}[thb]
\scriptsize
  \renewcommand{\arraystretch}{1.1}
  \centering
  \setlength{\tabcolsep}{0.2mm}{
  \caption{Comparisions of state-of-the-art detectors on COCO.}
  \label{tsota}
  \begin{tabular}{@{}l|c|c|c|ccccc@{}}
	\toprule
 	\textbf{Models}&	\textbf{Size}&	\textbf{Param.}(M)&	\textbf{FLOPs}(G)&	\textbf{AP}(\%)& ${\textbf{AP}}_{s}$(\%)& ${\textbf{AP}}_{m}$(\%)&  ${\textbf{AP}}_{l}$(\%)&  $\textbf{Latency}_{b=1}^{V100}$(ms) \\
	\midrule
	\multicolumn{9}{c}{\textbf{Transformers/CNNs} (non-real-time, pre-trained)}\\
	\midrule
	FPN\cite{fpn}&		-&			56.4&		145.8&			36.8&		17.5&		38.7&		47.8&	-\\
	BiFPN\cite{bifpn}&		1024&			21.0&			55.0&		49.3&		-&-&-&							42.8\\
	ASFF\cite{asff}&		800&			-&			-&		43.9&		27.0&		46.6&		53.4&			34.0\\
	PANet\cite{panet}&		-&			-&			-&		51.0&		32.6&		53.9&		64.8&			-\\
	RFPN\cite{rfpn}&		-&			-&			-&		51.3&		31.7&		54.6&		64.8&			-\\
	DETR-Def.\cite{ddetr}&	-&			40.0&		173.0&		43.8&		26.4&		47.1&		58.0&			-\\
	DETR-Dyn.\cite{dydetr}&	-&			-&			-&		47.2&		28.6&		49.3&		59.1&			-\\
	DETR-DND\cite{dddetr}&	1333&			48.0&		\textcolor{red}{265.0}&		48.6&		-&-&-&-\\
	DINO\cite{dino}&		1333&		47.0&		\textcolor{red}{860.0}&	51.0&		-&-&-&-\\			
	Swin-T\cite{swin}&		-&	\textcolor{red}{86.0}&\textcolor{red}{745.0}&		50.5&		-&-&-&-\\
	Swin-S\cite{swin}&		-&	\textcolor{red}{107.0}&\textcolor{red}{838.0}&	51.8&		-&-&-&-\\
	Swin-B\cite{swin}&		1280&	\textcolor{red}{145.0}&\textcolor{red}{982.0}&	51.9&		-&-&-&-\\
	\midrule
	\multicolumn{9}{c}{\textbf{CNNs} (real-time, without pre-trained)}\\
	\midrule
	Yolov4\cite{yolov4}&	608&		64.4&			142.8&	43.5&		26.7&		46.7&		53.3&		-\\
	YoloX\cite{yolox}&		640&		54.2&			155.6&		49.7&		-&-&-&						14.5\\
	PPYoloE\cite{ppyoloe}&	640&		52.2&			110.1&		50.9&		31.4&		55.3&		66.1&		12.8\\
	Yolov5\cite{yolov5}&	640&		46.5&			109.1&	49.0& 		-&-&-&						10.1\\
	Yolov6\cite{yolov6}&	640&		58.5&			144.0&	51.7& 		-&-&-&						-\\
	Yolov7\cite{yolov7}&	640&		36.9&			104.7&	51.2& 		31.8&		55.5&		65.0&		6.3\\
	Yolov8\cite{yolov8}&	640&		43.7&			165.2&	52.9& 		-&-&-&						-\\
\textbf{SYolo}(ours)&		640&		57.2&			142.5&	\textbf{53.1}& 	\textbf{35.5}&	 \textbf{58.7}&\textbf{69.9}& 			10.3\\
	\bottomrule
  \end{tabular}}
\end{table}
\section{Conclusion}
We explore the nature of representation learning and identify that the fusion of different level feature maps faces the issue of feature misalignment. To address this, we introduce the SNI. Additionally, we propose the ESD to mitigate spatial information loss during downsampling and further optimize lightweight convolutional methods for computational efficiency in lightweight models. Based on these advancements, we propose the SA solution for real-time detection, where all tested detectors surpass their original performance and achieve state-of-the-art results.

Object detection has significantly benefited from the FPN-like paradigm in past few years. However, recent developments indicate that newer technologies, including innovative architectures and training strategies, may offer superior performance for modern detectors, making the FPN-like paradigm potentially limiting. In many cases, the effectiveness of the FPN-like paradigm can be enhanced through the application of SNI to address feature misalignment issues. Fully addressing the challenge of misalignment during feature fusion in existing vision models requires further research efforts.
\begin{table}[h]
\scriptsize
  \renewcommand{\arraystretch}{1.0}
  \centering
  \setlength{\tabcolsep}{1.2mm}{
  \caption{Architecture of the SYOLO. The E-ELAN/C2f is frome the YOLOv7/8.}
  \label{tarchi}
  \begin{threeparttable}
  \begin{tabular}{@{}c|c|c|c|c@{}}
	\toprule
 	\textbf{Stage}&	\textbf{Block}-l/s/n$\ddagger$&	\textbf{Channels}-l/s/n$\ddagger$&	 \textbf{K. size}&	\textbf{Stride}	\\
	\midrule
	\multicolumn{5}{c}{\textbf{Backbone}}\\
	\midrule
	\textbf{P1}&	Conv&									64 /16 /16&		 	 3&2\\
	\midrule
	\textbf{P2}&	ESD-\uppercase\expandafter{\romannumeral1}/\uppercase\expandafter{\romannumeral2}&	128 /32 /32&			 3, 4&2\\
	&		C2f ×3/2/1&								128 /32 /32& 			 3, 1&1\\
	\midrule
	\textbf{P3}&	ESD-\uppercase\expandafter{\romannumeral1}/\uppercase\expandafter{\romannumeral2}&	256 /64 /64&			 3, 4&2\\
	&		C2f ×6/3/1&								256 /64 /64& 			 3, 1&1\\
	\midrule
	\textbf{P4}&	ESD-\uppercase\expandafter{\romannumeral1}/\uppercase\expandafter{\romannumeral2}&	512 /128 /128&			 3, 4&2\\
	&		C2f ×6/3/1&								512 /128 /128&			 3, 1&1\\
	\midrule
		&	ESD-\uppercase\expandafter{\romannumeral1}/\uppercase\expandafter{\romannumeral2}&	1024 /256 /256&			 3, 4&2\\
	\textbf{P5}&	C2f ×3/2/1&								1024 /256 /256&			 3, 1&1\\
		&	SPP ×1&									1024 /256 /256& 			 -&-\\
	\midrule
	\multicolumn{5}{c}{\textbf{Neck}(SA)}\\
	\midrule	
		&	 SNI&									1536 /384 /284&			 -&-\\
	\textbf{P5-P4}&	 E-ELAN$\dagger$×1&							2048 /512 /512&			 3, 1&1\\
	\midrule
		&	 SNI&									512 /192 /192&			 -&-\\
	\textbf{P4-P3}&	 E-ELAN$\dagger$×2/2/1&							512 /256 /256&			 3, 1&1\\
	\midrule
		&	ESD-\uppercase\expandafter{\romannumeral1}/\uppercase\expandafter{\romannumeral2}&	 256 /128 /128&			 3, 4&2\\
	\textbf{P3-P4}&	E-ELAN$\dagger$×2/2/1&							 1024 /512 /512&			 3, 1&1\\	
	\midrule
		&	ESD-\uppercase\expandafter{\romannumeral1}/\uppercase\expandafter{\romannumeral2}&	 512 /256 /256&			 3, 4&2\\
	\textbf{P4-P5}&	E-ELAN$\dagger$×2/2/1&	 						 2048 /1024 /1024&			 3, 1&1\\
	\bottomrule
  \end{tabular}
  \begin{tablenotes}
\scriptsize
  \item[$\dagger$] The first 3×3 vanilla convolutional layer of the E-ELAN is replaced by the GSConvE-\uppercase\expandafter{\romannumeral1}, and the end pointwise convolutional layer of the E-ELAN is replaced by the C2f. 
  \item[$\ddagger$] The 'l/s/n' means the model scale of the large/samll/nano. 
  \end{tablenotes} 
  \end{threeparttable}}
\end{table}
\clearpage  
\section*{Acknowledgements}
This work is supported by National Natural Science Foundation of China (Grant No. 52172381), Graduate Research Innovation Foundation of Chongqing Jiaotong University (Grant No. CYB240259). The author would also like to thank Prof. Dr. Jun Li, Prof. Dr. Qiliang Ren, and Ms. Xinxin Liu (Intelligent Transportation Big Data Center of Chongqing Jiaotong University) for providing GPUs.
%
%
\bibliographystyle{splncs04}
\bibliography{reference}
\end{document}